# Implementing a Portable Clinical NLP System with a Common Data Model – a Lisp Perspective

Yuan Luo*
Dept. of Preventive Medicine
Northwestern University
Chicago, USA
yuan.luo@northwestern.edu

Peter Szolovits*
CSAIL
MIT
Cambridge, USA
psz@mit.edu

*Abstract*— This paper presents a Lisp architecture for a portable NLP system, termed LAPNLP, for processing clinical notes. LAPNLP integrates multiple standard, customized and in-house developed NLP tools. Our system facilitates portability across different institutions and data systems by incorporating an enriched Common Data Model (CDM) to standardize necessary data elements. It utilizes UMLS to perform domain adaptation when integrating generic domain NLP tools. It also features stand-off annotations that are specified by positional reference to the original document. We built an interval tree based search engine to efficiently query and retrieve the stand-off annotations by specifying positional requirements. We also developed a utility to convert an inline annotation format to stand-off annotations to enable the reuse of clinical text datasets with inline annotations. We experimented with our system on several NLP facilitated tasks including computational phenotyping for lymphoma patients and semantic relation extraction for clinical notes. These experiments showcased the broader applicability and utility of LAPNLP.

## I. INTRODUCTION

Much of the content of medical records is locked up today in narrative text. There are many examples where we would greatly benefit from access to the facts, observations and plans encoded in clinical narratives [1, 2]. Any analysis process, whether focused on quality improvement or clinical research, needs a systematic representation of these data. Furthermore, we need to develop the ability to measure phenotype and environment with an efficiency approaching that of our abilities to use high-throughput biological techniques to measure genotype and related molecular biological phenomena such as the transcriptome, proteome, etc. Clinical records provide the best source of phenotypic and environmental information about human beings, in whom we cannot ethically manipulate environment or phenotype simply to satisfy our scientific curiosity.

For medical natural language processing, we still operate in the mode that every research group develops and applies its own new text analysis methods to specific data sets of interest. Furthermore, the data sets and the analysis results have been difficult to share because of concerns over residual privacy-threatening content. Finally, with rare exceptions, most of the tools that have been developed and made public for various components of text analysis operate as independent processes that transform an input document into a different output document to which have been added various textual annotations, often in idiosyncratic representations. This makes it quite difficult to chain together sequences of operations. Although several recent projects have achieved reasonable success in analyzing certain types of clinical narratives [3-6], efforts towards a common data model (CDM) to ensure portability have only begun. For example, the Observational Medical Outcomes Partnership (OMOP) CDM currently only has NOTE and NOTE_NLP tables and is in need of extension and refinement. In addition, representing the NLP extracted information using graphs to mine their relations and interactions is a growing need [7] yet unmet by current CDMs.

In this paper, we build a Lisp Architecture for Portable NLP (LAPNLP) for clinical notes. LAPNLP annotates clinical notes using both in-house developed NLP tools and standard and customized tools that are integrated using an automatic wrapper generation utility we built. LAPNLP stores the derived annotations into a relational database using the schemas defined in a rich Common Data Model (CDM). The rich CDM we adopted subsumes the OMOP CDM and defines related tables for graph representation and manipulation. We make LAPNLP code available at https://github.com/yuanluo/lapnlp.

## II. LAPNLP COMMON DATA MODEL

LAPNLP CDM's fundamental organization is around the following key ideas:

(1) Data structures represent documents, corpora and annotations. Each of these is persistent in a relational database, so that processing can be performed incrementally, and previous annotations can always be re-used in subsequent processes. Furthermore, we provide extended CDM tables that are designed for graph representation and mining in order to meet the need for semantic relation and network analysis.

(2) LAPNLP uses stand-off annotations, whereby the original document is never altered or directly marked up by processing steps. This eliminates the need to maintain multiple versions of a document depending on what mark-ups have been made on it, and it eliminates the need to re-parse those files into an internal representation that support the next processing step. We also built an in-memory interval tree data structure to facilitate the efficient query and retrieval of stand-off annotations.

(3) LAPNLP is based on the Lisp programming language, which is a highly expressive language, where functions are first-class objects, encouraging meta-programming.

* indicates equal contribution. This study was supported in part by grant R21LM012618-01 from the National Library of Medicine of the NIH.

LAPNLP implements an automated interface system to C, C++, and Java programs, as well as to MySQL and Postgres databases. This enables the integration of in-house built, customized and standard NLP tools as well as knowledge-database-supported domain adaptation within LAPNLP.

*A. Core Tables in the LAPNLP Common Data Model*

In LAPNLP we use mechanisms akin to marshalling complex data into a form that can be stored in a relational database and unmarshalling them into memory structures when data are retrieved from the database. This facility is important first for the obvious reason that LAPNLP's in-memory data structures are evanescent and need to be saved and restored from a persistent store. We consider the corpora, documents and annotations to be the primary resource in any LAPNLP application, and want to assure that these are available to other teams doing development and analysis on the same data, perhaps using other NLP frameworks. Relational databases with mutually agreed schemas are a common way towards portability, and is our design choice here.

Figure 1 shows the core relational tables in the LAPNLP CDM. In an object-oriented language such as Lisp, it is quite common for a class's subclasses to define additional fields or attributes beyond those present in the original class. Yet relational databases define tables with a fixed column structure, which makes a direct mapping from fields to columns problematic: one might need distinct tables to store the instances of every subclass separately. The triple-store alternatives such as Resource Description Framework (RDF), require many SQL operations to store an individual instance and many self-joins of potentially huge tables to reconstruct one. This dilemma, and previous examples such as OMOP CDM, have led us to adopt a hybrid strategy. In this, a small number of tables store the columns that correspond to those fields or attributes in the classes we use that are most commonly used and for which efficient retrieval by value (hence indexing) is most needed. Such core tables include the corpora, documents and annotations tables, type definition tables, as well as graph representation tables. Other varying attributes in these core tables are serialized to an additional text field from which they can easily be reconstructed.

In the relational implementation, we maintain a corpora table that holds the name and unique id of the corpus, meta-data, and the above-mentioned structural description. A corpora-documents table holds the n-n association between corpora and documents based on their unique id's. We allow many-to-many mappings as documents can be reused for multiple corpora in different tasks. The documents table holds the name and unique id of each document, associated meta-data, and its content as a UTF-8 long text field. When a document is to be processed, the relational database representation of its annotations is unmarshalled into an in-memory interval tree that is used extensively as analysis programs search for existing annotations, add new ones, or update existing ones. At checkpoints or when processing of that document is completed, these annotations are updated in the persistent store.

Each annotation contains one identified data field, which contains the principal value of the annotation. For example, a tagger may produce an unambiguous preferred part of speech for a token. This would be recorded by creating an annotation for the same span of text, and placing the tag as its data value. These values are indexed in the persistent store, and thus retrieval by value is quite efficient. Specific annotations normally have additional fields, which are serialized to a text field. For example, the "data" column in the annotations table store the text content that are annotation type dependent: A UMLS CUI annotation maintains the CUI information.

We additionally introduce the notion and table of instances. Instance is a machine learning concept referring to a sample that is task dependent and can be a document (e.g., document classification), a pair of named entities (e.g., relation classification), or a set of documents (e.g., computational phenotyping based on all clinical notes of a patient during an encounter), etc. Such set relationships are stated in the table "instances_content". The table "instance_sets" can be used to store training, validation and testing datasets associated with any corpora. The "groundtruth" table is then used to store the gold-standard with respect to corresponding datasets.

*B. Storing and Manipulating Graph Structures*

Graph mining has become increasingly important in clinical NLP in recent years [7]. For example, dependency graphs from syntactic parses of sentences can be used to assist tasks such as semantic role labeling via graph convolutional networks [8]. Medical concepts connected by semantic relations form a knowledge graph that can guide the reasoning towards drug-drug interactions and adverse drug events, among others [9, 10].

The graph representation is the foundation for graph mining, and along with upstream steps including direct regular expression feature extraction, leads to the generation of semantically and syntactically enriched features. These features then support either rule based, feature vector space based or kernel based relation extraction systems. Thus we also include graph related tables among LAPNLP core tables. The graph table stores all the graphs' ids, names and types. The detailed structure of each individual graph is stored in the "linkage_graph" table using the node-edge list format [11]. We provide CDM support for significant graph mining such as frequent subgraph mining [11]. Thus we store the mined significant graphs in the "sig_subgraph" table and use the "lg_sigsub" table to store the matching and mapping information between entire graphs ("linkage_graph") and significant subgraphs ("sig_subgraph").

III. Efficient Handling of LAPNLP Annotations

Even a document of modest size may accumulate thousands to hundreds of thousands of different annotations. For example, each word may be annotated multiple times on its orthography, parts of speech, stemmed root, etc. Each phrase may have multiple annotations as well, including phrase types, semantic categories, presence in particular dictionaries, etc. Moreover, annotations may overlap, for example, "congenital defect of the heart", "congenital", "congenital defect", "heart", "defect of the heart" may each get separate annotations. Consequently, it is important to be able to support queries based on positional relationships between annotations efficiently.

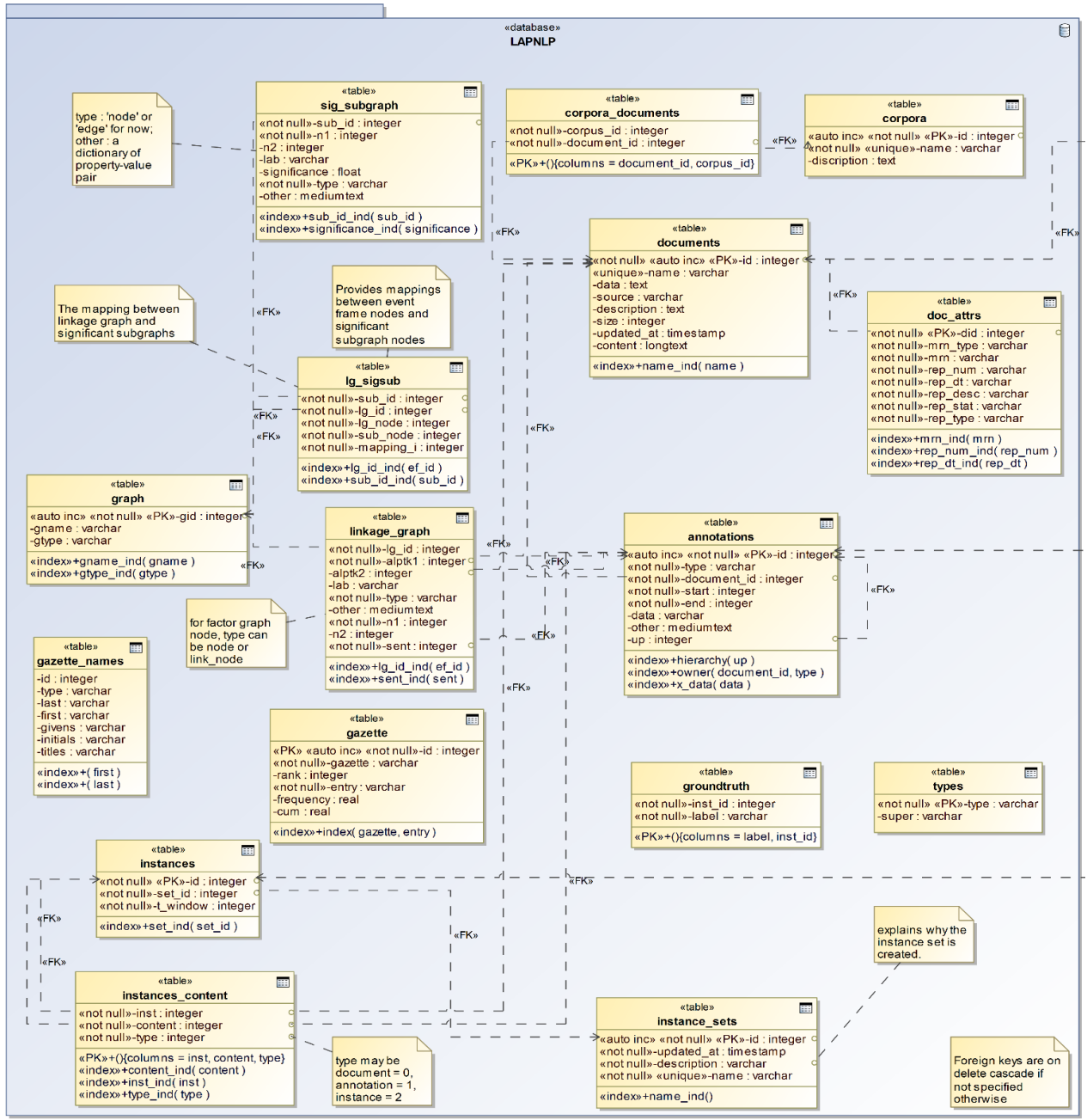

Figure 1 Core Tables in LAPNLP Common Data Model. FK: foreign key; PK: primary key.

*A. Annotations in Stand-off Format*

Many early NLP systems and corpora used in-line annotations where annotations are marked and embedded in the text (e.g., XML style markups). In comparison, a stand-off annotation identifies the starting and ending position of the text to which it applies, and contains various types of additional information specific to the type of annotation. Stand-off annotations have the benefit that the original document is never altered or directly marked up by processing steps, and is adopted by many modern NLP systems [3-6]. LAPNLP adopts stand-off annotations and stores them in an interval tree data structure that makes annotation retrieval fast and management efficient. Stand-off annotations also enable the user to store the key components of rule-based systems (e.g., regular expression matches) in the format of CDM, thus enabling the reuse, adaptation and extension of many existing rule-based clinical NLP systems (e.g., NegEx [12]).

*B. Building the Interval Tree for Stand-off Annotations*

With stand-off annotations, queries will require efficient retrieval of annotations based on their spatial relation to another annotation. For example, a program may need to extract the two tokens successively following a given token to help with feature extraction. As Allen has shown for intervals, there are thirteen mutually exclusive possible relations between two intervals [13]. If we consider two intervals, $i$ and $j$, with starting and ending positions $i_s$ and $i_e$, and $j_s$ and $j_e$, respectively, they may have any of the 13 relationships shown in TABLE 1. In the degenerate case where some interval is null (i.e., $i_s = i_e$), then it is possible for more than one of the above relations to be satisfied. For example, given a null $i$ whose start and end coincide with a non-null $j$'s $j_s$, we could say that $i\ m\ j$ or that $i\ s\ j$.

TABLE 1 POSSIBLE RELATIONS AMONG INTERVALS, IN ANALOGY WITH ALLEN'S TEMPORAL LOGIC [13]. IN OUR NOTATION, $i$ AND $j$ ARE INTERVALS, AND $i_s$ IS THE START POSITION AND $i_e$ THE END OF AN INTERVAL. ACRONYM: M – MEETS, D – DURING, S – STARTS, F – FINISHES, O – OVERLAPS.

| Symbol | Definition | Inverse |
|---|---|---|
| $=$ | $i_s = j_s \wedge i_e = j_e$ | |
| $<, >$ | $i_e < j_s$ | $i_s > j_e$ |
| $m, mi$ | $i_e = j_s$ | $j_e = i_s$ |
| $d, di$ | $i_s > j_s \wedge i_e < j_e$ | $j_s > i_s \wedge j_e < i_e$ |
| $s, si$ | $i_s = j_s \wedge i_e < j_e$ | $i_s = j_s \wedge i_e > j_e$ |
| $f, fi$ | $j_s < i_s \wedge i_e = j_e$ | $j_s > i_s \wedge i_e = j_e$ |
| $o, oi$ | $i_s < j_s < i_e < j_e$ | $j_s < i_s < j_e < i_e$ |

To support efficient retrieval of all annotations of a certain type that satisfy a specific relation to a query annotation, we load all annotations of a document to an interval tree in memory. We use a canonical ordering on intervals such that they are in ascending order first by start position and then, if those are equal, by end position. We build the interval tree as a data structure augmented from the red-black tree that stores all the intervals in this canonical order [14]. The augmentation is such that each node maintains the minimum and the maximum end positions of the intervals stored under that node. This information changes only on insertions into or deletions from an interval tree, and imposes no significant query cost. We perform efficient searches for any of the Allen's interval algebra relations by converting them to a base relation. For example, if we seek all intervals $i$'s that have the $o$ relation to the query interval $b$, consider what we need to do when searching a node of the interval tree where a certain interval $x$ is stored. First, we check whether $x$ satisfies $o$ to $b$, and add it to the answer if so. Then, we see whether any of the intervals in the left or right branches at this node might include other intervals that satisfy the $o$ relationship. Clearly, if the maximum end of all intervals in the left branch is not larger than $b_s$, then we need not search that branch. This is the fundamental intuition why interval tree data structure enables fast query through pruning branches of the search path.

IV. PROCESS FLOW OF LAPNLP

The LAPNLP process flow can be divided into several key components including: finding structures in documents (i.e., breaking documents into sections), breaking documents into sentences, breaking sentences into tokens, tagging tokens with part-of-speech, grouping tokens into phrases (chunking), deep parsing, identifying UMLS annotations on tokens and phrases, recognizing assertions such as negations, generating features and various machine learning components. In this section, we describe mechanisms and features for individual components.

*A. In-house Built Tool for Section Detection*

In clinical NLP, it is an important process to annotate the section and hierarchic subsection portions of a document as section name may indicate the meaningful nature of the recorded clinical information (e.g., past medical history) [15]. In order to guide the detection of the section structure of the clinical notes, we store an XML representation of the overall "grammar" of the documents in the corpus. For example, a discharge summary may contain sections corresponding to the patient's past medical history, family and social history, medications, physical examination, tests and results, procedures, and disposition. These sections can usually be reliably detected by identifying their corresponding section headings as the section start and the following section headings as (right after) the section end. In addition, clinical notes are often based on templates that are usually specific to hospitals and institutions that render them. Thus we allow the users to add to the XML representation the patterns that describe an entire template such as the following excerpt from a pathology report: "Differential: ____ % polys, ____ % bands, ____ % lymphs, ____ % monos, ____ % eos, ___ % basos, ___ % blasts."

Both types of patterns are specified in the form of an XML guideline. Of the elements of an XML expression, the most critical one is the pattern element, which specifies how to recognize a section or section head. We give an XML head the option to define a property list, where properties usually define regular expression matching options to capture section headings or section bodies, and section naming convention.

To allow capture of more structured information of a section, we provide mechanisms to specify attributes of certain section. The attributes are often constructed from the matching groups of the regular expression patterns. Some of the structured sections contain subsections that are structured sections, and our algorithm can recursively find those subsections, provided that the XML guideline specifies the candidate subsections to search for.

*B. In-house built Tool for UMLS Annotation*

The Unified Medical Language System (UMLS) is aimed at building a warehouse that links together many health and biomedical vocabularies, codes and standards. We leverage UMLS for domain adaptation of the generic NLP tools incorporated by LAPNLP, as well as for in-house developed NLP tools for tasks such as named entity recognition.

For named entity recognition, we provide a utility that enables a user to annotate a section of text with UMLS annotations, which includes:

- CUI annotation: A Concept Unique Identifier (CUI) corresponds to a unique concept.
- TUI annotation: A TUI is a semantic type ID for one of the semantic categories in the UMLS Semantic Network.
- SP-POS annotation: Annotation on individual tokens showing (one of) the Specialist lexicon parts of speech possible for that token.

LAPNLP annotates a sentence with UMLS annotations using the following procedure. We investigate each of the $n \times (n-1)$ subsequences of tokens in a sentence and look it up in the UMLS Metathesaurus. We then perform a greedy search to find the longest token subsequences with a match-ing UMLS concept unique identifier (CUI). Employed heuristics to guide the greedy search include ignoring case in matching, eliminating subsequences that are fully contained in longer sequences, eliminating interpretations of single tokens that fall into function-word grammatical categories, and ignoring punctuation. After that, we look up multiple UMLS mapping tables and obtain other annotations from CUIs.

*C. Domain Adaptation of Existing NLP Tools*

After the section detection step, the following steps are frequently performed by multiple generic and domain specific NLP tools. For example, sentence breaking as implemented in OpenNLP, stemming as implemented in Porter Stemmer [16], POS tagging as in the Link Grammar Parser [17], and parsing as in Stanford Parser [18]. The examples above showcased the availability of multiple existing tools for well-established NLP steps. Thus we provide an automated wrapper generation utility to integrate these tools under their corresponding language environment (e.g., C for Link Grammar Parser; Java for OpenNLP and Stanford Parser). Please see our github repository for details.

*D. Converting Sentence Dependency to Graph Structures*

In NLP, the syntactic structure of a statement often corresponds at least approximately to the ways in which the semantic parts may be combined to aggregate the meaning of the overall statement [18]. The two-phase sentence parsing described above produces the dependency linkage structure of a sentence. We translate dependency parses to a graph representation of the relations, where the nodes are concept annotations and the edges are syntactic dependencies among the concepts.

V. USE CASES OF LAPNLP

In this section, we describe several use cases of LAPNLP in processing different types of clinical notes for tasks including computational phenotyping (an example of long text understanding), semantic relation extraction (an example of short text understanding) and converting inline annotations to stand-off annotations.

*A. Computational Phenotyping of Lymphoma*

In Luo et al. [19, 20], we described the application of LAPNLP to pathology reports in order to automatically classify the lymphoma subtypes. One specific novel application of LAPNLP is the unsupervised extraction of relations from narrative text and our use of these relations as features for further analysis. The intuition, in the context of classifying lymphoma subtypes based on pathology reports, is that one relation such as "[large atypical cells] express [CD30]" leads to a certain belief that the patient might have Hodgkin lymphoma. Adding "[large atypical cells] have [Reed-Sternberg appearance]" increases the belief in Hodgkin lymphoma. Thus we used two-phase parsing to obtain domain refined dependency parsing and construct the sentence graphs from the parses. We further mined frequent subgraphs as relation features and stored all parses, graphs and subgraphs (relations) in the CDM tables of LAPNLP. Our system achieved the state-of-the-art performance when representative lymphoma subtypes classification (>0.8 F1) and clustering (>0.7 F1) and feature analysis identified subgraph features that contribute to improved performance, which are consistent with the current knowledge about lymphoma classification.

*B. Semantic Relation Extraction*

Biomedical relation extraction is critical in understanding clinical notes, facilitating automated diagnostic reasoning and clinical decision making. We applied LAPNLP on the relation classification dataset from the i2b2/VA challenge, which contains the relations that may hold between medical problems and treatments, between medical problems and tests, as well as between pairs of medical problems [21]. Because the two medical concepts each span multiple words in the original sentence, they can be represented as intervals with start and end positions counted by number of words. In particular, we observe the need to distinguish the concepts vs. context text, and further differentiating the context text into text preceding the first concept, between the concepts, and succeeding the second concept [22]. The interval tree utility from LAPNLP provides a handy tool for extracting features from individual segments subject to various positional requirements. The stand-off annotation design of LAPNLP makes it easier to extract multiple types of features (e.g., words, other concepts, POS-tags) that meet the positional requirements. In this aspect, we used LAPNLP to explore different feature configurations and built the foundation for several neural network models to tackle the relation classification problem [23, 24].

*C. Converting Inline Annotation to Stand-off Annotation*

Some of the community shared-task challenges have their annotations use the in-line format. However, the stand-off annotation format is adopted by most modern clinical NLP pipelines including cTAKES [3], MetaMap [5], Leo [4] and CLAMP [6]. To facilitate the continued usage of many earlier released shared-task challenge dataset, we provide a utility to convert inline annotation to stand-off annotation. Figure 2 shows an example of such conversion using a sentence with private health information (PHI). To import in-line annotation datasets in a form consistent with standoff annotations, we read in the XML file with the option to preserve blanks, break up the data to create documents out of each individual record, and then concatenate all the actual text in the TEXT elements, stripping out the PHI markers. However, we note the location of the PHI tags and use these as the start and end positions for our standoff annotation.

In-line annotation

The$_3$ patient$_{11}$ underwent$_{21}$ an$_{24}$ ECHO$_{29}$ and$_{33}$ endoscopy$_{43}$ at$_{46}$ <PHI TYPE="Hospital">Beth$_{51}$ Israel$_{58}$ Deaconess$_{68}$ Medical$_{76}$ Center$_{83}$</PHI> on$_{86}$ <PHI TYPE="Date">April$_{92}$ 28$_{95}$</PHI>.

Stand-off annotation

| Start | End | Annotation Type | Annotation Attribute |
|---|---|---|---|
| 48 | 83 | PHI | Type=Hospital |
| 88 | 95 | PHI | Type=Date |
| … | … | … | … |

Figure 2 Comparison of in-line annotation and stand-off annotation (counted in characters). PHI: Private Health Information.

VI. DISCUSSIONS

Variation in data models, coding systems, and narrative styles used at different institutions make it difficult to conduct large-scale analyses of observational healthcare databases. Our system is a step to address the problems of analyzing narrative text from such data and enhances its portability by utilizing an enriched CDM and its standardized terminologies. We harmonized the output from multiple standard, customized, and in-house developed NLP tools into the CDM for researchers to conduct systematic analysis at a larger scale. However, we note that our CDM has many additional components than other CDMs currently in use at research consortia such as OMOP CDM. For example, we have a data column for extended annotation attributes and several database tables related to graph representation and mining. These additional components represent our own perspectives towards what is necessary in a clinical CDM. We expect more use cases and a formal utility study is needed to validate their usefulness.

We have not yet perfected either the individual components of language processing systems or the pipelines whereby they can be strung together to achieve specific language processing tasks, LAPNLP is an environment intended for researchers who are investigating improving such methods rather than end-users who wish to apply them to specific clinical data sets. For this latter application, LAPNLP shares with other such integrated frameworks [3-6] the unfortunate difficulties of any system that is assembled from many components—namely, the need to install, configure and tie together all the components. We consider this an acceptable burden for developers, who should possess the requisite computing skills to do so, but it lies generally beyond the abilities of end users.

## VII. CONCLUSIONS

We developed LAPNLP, a portable phenotyping system that is capable of integrating standard, customized and in-house developed NLP tools. LAPNLP is based on an enriched CDM, and stores annotations from both automated NLP tools and rule-based systems (e.g., regular expression matches) using stand-off annotation format, in order to standardize necessary data elements and ensure portability. Comparing to file system based pipelines such as UIMA CAS stacks and BioC, the LAPNLP CDM uses a database as the persistent store and has the advantages offered by database management systems. Compared with OMOP CDM, LAPNLP CDM is extended into 20 carefully designed tables including tables for graph representation and mining. LAPNLP also built an interval tree to support fast and efficient queries and retrieval of stand-off annotations. We showcased the utility and benefits of LAPNLP through use cases including computational phenotyping on long clinical text and relation extraction on short text.